\documentclass[10pt,conference,a4paper]{IEEEtran}
\usepackage[bottom=1.8cm, top=1.8cm, left=1.6cm, right=1.6cm]{geometry}
\usepackage{cite}
\usepackage{url} 
\usepackage{hyperref}
\hypersetup{
    colorlinks=true,
    linkcolor=black, 
    citecolor=black, 
    filecolor=black, 
    urlcolor=black   
}
\usepackage{xcolor}
\usepackage{amsmath,amssymb,amsfonts}
\usepackage{algorithm}
\usepackage{algpseudocode}
\usepackage{graphicx,dblfloatfix}
\usepackage{textcomp}
\usepackage{eqparbox}
\usepackage{float}
\usepackage{multirow}
\usepackage{subcaption}
\usepackage{caption}
\usepackage[font=small,labelfont=bf]{caption}
\newcolumntype{M}[1]{>{\centering\arraybackslash}m{#1}}

\def\BibTeX{{\rm B\kern-.05em{\sc i\kern-.025em b}\kern-.08em
    T\kern-.1667em\lower.7ex\hbox{E}\kern-.125emX}}

\begin{document}

\title{Parallel Neural Computing for Scene Understanding from LiDAR Perception in Autonomous Racing}


\author{
\IEEEauthorblockN{Suwesh Prasad Sah}
    \IEEEauthorblockA{\textit{\small Dept. Computer Science and Engineering} \\
    \textit{\small National Institute of Technology}\\
    \small Rourkela, India\\
    \small suwesh081@gmail.com}
    
\and
    \IEEEauthorblockN{Suchismita Chinara}
    \IEEEauthorblockA{\textit{\small Dept. Computer Science and Engineering} \\
    \textit{\small National Institute of Technology}\\
    \small Rourkela, India\\
    \small suchismita@nitrkl.ac.in}
}

\maketitle

\begin{abstract}
Autonomous driving in high-speed racing, as opposed to urban environments, presents significant challenges in scene understanding due to rapid changes in the track environment. Traditional sequential network approaches may struggle to meet the real-time knowledge and decision-making demands of an autonomous agent covering large displacements in a short time. This paper proposes a novel baseline architecture for developing sophisticated models capable of true hardware-enabled parallelism, achieving neural processing speeds that mirror the agent's high velocity. The proposed model (Parallel Perception Network (PPN)) consists of two independent neural networks, segmentation and reconstruction networks, running parallelly on separate accelerated hardware. The model takes raw 3D point cloud data from the LiDAR sensor as input and converts it into a 2D Bird's Eye View Map on both devices. Each network independently extracts its input features along space and time dimensions and produces outputs parallelly. The proposed method's model is trained on a system with two NVIDIA T4 GPUs, using a combination of loss functions, including edge preservation, and demonstrates a 2x speedup in model inference time compared to a sequential configuration. Implementation is available at: \href{https://github.com/suwesh/Parallel-Perception-Network}{\textcolor{purple}{github/ParallelPerceptionNetwork}}. Learned parameters of the trained networks are provided at: \href{https://huggingface.co/suwesh/ParallelPerceptionNetwork}{\textcolor{orange}{huggingface/ParallelPerceptionNetwork}}.
\end{abstract}


\section{Introduction}
Autonomous racing promises to deliver safer and more reliable self-driving vehicle technology by pushing the limits of autonomous driving. Understanding scene dynamics is crucial for autonomous driving, but the challenge intensifies in racing as these vehicles move at higher velocities. Faster speeds necessitate quicker perception and decision-making. These challenges can be addressed if the autonomous agent can perceive and understand its environment to perform multiple tasks in a single model inference cycle.

In recent years, simulators have been primarily focused on research for autonomous vehicles \cite{dosovitskiy2017carla} \cite{shah2018airsim} \cite{wymann2000torcs}. However, because simulators provide noiseless sensor data to models, they struggle during real-world deployments. This paper departs from simulator-based methods and uses LiDAR sensor data from the recently released RACECAR dataset \cite{kulkarni2023racecar}, the first open dataset for full-scale and high-speed autonomous racing.

\begin{figure}[t]
    \centering
    \includegraphics[width=1\linewidth]{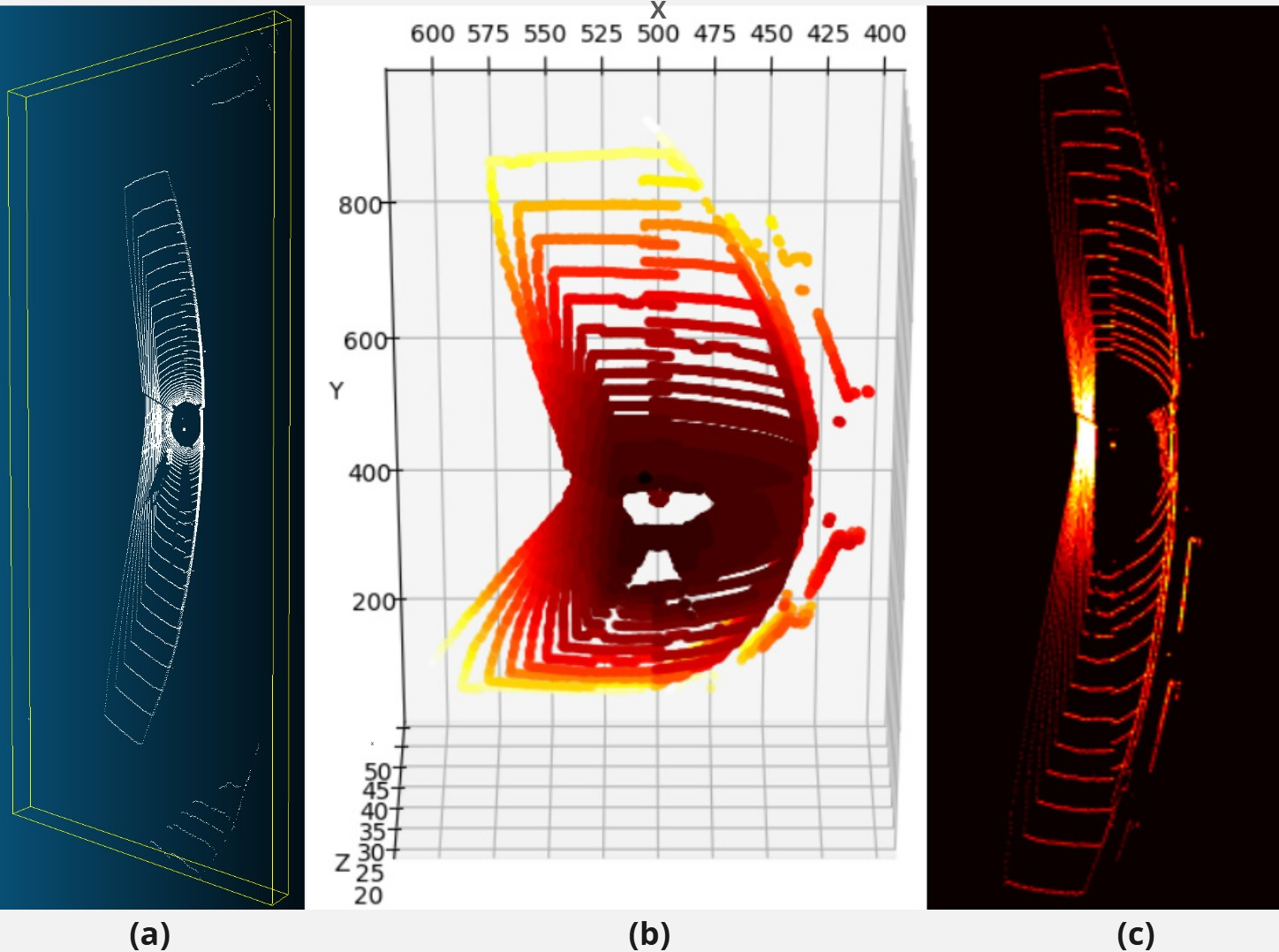}
    \caption{Conversion of 3D point clouds into 2D BEV map. This process involves: (a) Point clouds in 3D space. (b) Voxelization, where the 3D space is divided into discrete voxels and each voxel holds the max z-axis value. (c) 2D BEV map obtained by projecting 3D voxels onto a 2D plane by taking the maximum along z-axis.}
    \label{fig:PointCloudData}
\end{figure}

The dataset contains rich point cloud data that provides a detailed 3D representation of the world around the autonomous agent. These point clouds are a geometric data structure representing the spatial arrangement of points in a 3D space. RACECAR’s point clouds collect x, y, and z coordinates and intensity values for each laser scan, covering a $360^\circ $ field of view with a range of $120m$ as can be observed in Fig \ref{fig:PointCloudData}. Since point clouds struggle to capture spatial relationships directly, the LiDAR scans in nuScenes \cite{caesar2020nuscenes} format are converted into 2D Bird’s Eye View (BEV) maps \cite{chang2024bevmap} to efficiently capture the scene layout and features. A sequence of these scans is stacked along the time dimension, allowing our approach to learn a short history of the environment's state.

This paper proposes a parallel neural network computing baseline with a deep learning model that can perform segmentation along the space-time dimension and scene reconstruction in a single model inference. The core architecture of networks in the PPN model is an encoder-decoder convolutional neural network, drawing inspiration from successful architectures like UNet \cite{ronneberger2015u}, MotionNet \cite{wu2020motionnet}, and FPN \cite{lin2017feature}.

The two networks are a segmentation network with skip connections and a reconstruction network without skip connections. The skip connections capture high-level and low-level features while encoding and forward them to corresponding layers while decoding, this allows the segmentation network to accurately segment the input sequence of scenes providing the network with a brief history of travel. The training for PPN employs a combination of SmoothL1 loss and MSE loss with Canny Edge detection as a loss function, building upon the work proposed in \cite{pandey2018msce}. The loss function combination minimises absolute and squared differences between predicted and ground truth scenes. Adding edge preservation ensures that sharp features such as track boundaries are preserved.

GPUs are utilized as hardware accelerators \cite{matsuoka2009gpu} for high-performance computing while training and inference of deep neural networks due to advancements in GPU computing and performance, particularly in NVIDIA’s CUDA platform \cite{dehal2018gpu}. See Fig \ref{fig:parallelaccHW}. Hence, utilizing separate accelerated hardware for true hardware-enabled parallel computing mitigates the problems of latency in real-time perception for multiple tasks in autonomous high-speed racing, as it can be observed in the results section indicating a faster model inference time of the parallel configuration compared to a sequential one.

Finally, the rest of the paper is structured as follows. Section 2 is a review of related works. Section 3 elaborates on system methodology. Section 4 explains the experimental setup, results and analysis, performance evaluation, and comprehensive comparison. Finally, Section 5 is the conclusion.
\section{Related Work}
This section elaborates on related works on self-driving cars for racing and in urban areas using different approaches.

The authors in \cite{imamura2021expert} \cite{weiss2020deepracing} \cite{loiacono2010learning}  have provided end-to-end research in autonomous car racing, achieving high-level performance using a realistic simulator.
The RACECAR dataset is a pivotal contribution in \cite{kulkarni2023racecar}, providing rich multi-modular sensor data collected from fully autonomous Indy race cars, which can be utilized and analyzed for better evaluations in autonomous racing.
In paper\cite{wu2020motionnet}, a deep neural model is proposed called MotionNet, which stands out for its joint perception and motion prediction using 2D convolutions in a pyramid network instead of using 3D convolutions on a sequence of BEV maps.
The InsMOS approach in paper \cite{wang2023insmos} further advances the segmentation of moving objects in 3D LiDAR data by integrating instance information in the vanilla MotionNet architecture.
The authors in \cite{sun2019predicting} have focused on the future of instance segmentation, proposing a Contextual Pyramid ConvLSTM architecture to predict the evolution of the scenes. This approach has a computational overhead of RNN structures. While paper \cite{luc2018predicting} has applied Mask R-CNN to predict future instance segmentation.
Paper \cite{yang2018scene} has proposed frameworks on CNNs with a feature map-based approach where deconvolutions recover the feature maps to extract features that contribute to understanding driving scenes.
The authors in \cite{cui2021lookout} have presented a novel system that perceives the environment and predicts a diverse set of possible futures. The backbone of this framework is a CNN, which takes as input a history of LiDAR sweeps in the BEV map.
In paper, \cite{hu2020probabilistic}, a deep learning architecture that learns spatio-temporal representations through convolution modules is presented, which are decoded for future semantic segmentation.
Authors of \cite{kim2021stfp} have presented forecasting of traffic scenes using four modules utilizing 2D, 3D-CNN and Conv-LSTMs on a similar map representation approach.
Paper \cite{muhammad2022vision} shows various types of segmentation network architectures, like encoder-decoder based on convolution layers in scene understanding for autonomous vehicles.
\section{Methodology}
\begin{figure}[t]
    \centering
    \includegraphics[width=1\linewidth]{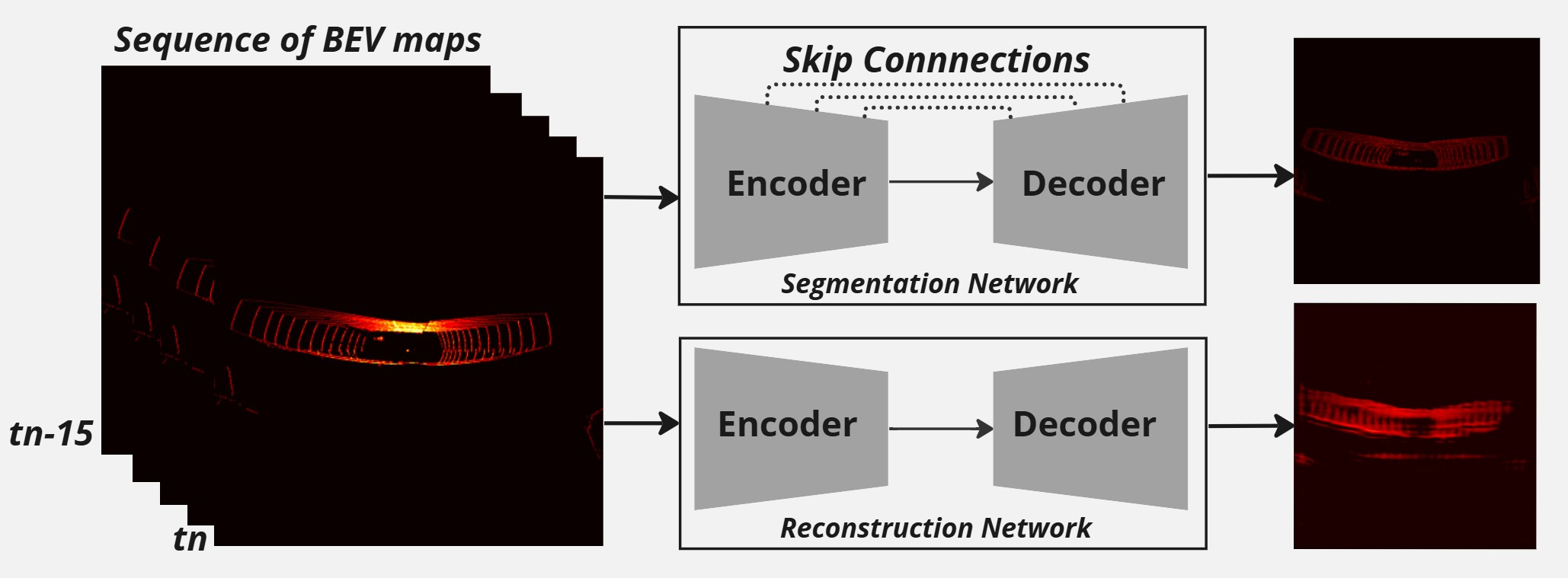}
    \caption{Overview of PPN. The segmentation network with skip connections is a spatio-temporal pyramid network, and the reconstruction network is an autoencoder.}
    \label{fig:overview of PPN}
\end{figure}
This section elaborates on system methodology, see Fig \ref{fig:overview of PPN}. Initially, the point clouds are extracted from LiDAR sweeps using the nuscenes-devkit, capturing [x, y, z, intensity] values for each point. To convert these 3D point clouds into a 2D Bird’s Eye View (BEV) map, a structured process is followed detailed in Algorithm \ref{algo1}. This process begins with voxelization, where the 3D space is divided into a grid of voxels. Within each voxel, max pooling is applied to the z-axis values to retain the highest elevation feature, effectively compressing the 3D structure into a 2D representation (See Fig \ref{fig:PointCloudData}).

Next, the pooled z-axis values are rescaled to a range of $0$ to $1$, ensuring standardized data for further processing. This rescaling is crucial for generating a binary BEV map, where a threshold distinguishes between occupied value of $1$ and free space value of $0$. The binary conversion facilitates clear interpretation by highlighting the presence or absence of objects within the mapped environment. This methodology, from voxelization to binary conversion, transforms raw LiDAR data into an insightful 2D representation, preserving critical elevation information and enhancing spatial analysis.
\begin{algorithm}
\caption{Convert 3D point clouds into 2D binary BEV map.}
\begin{algorithmic}[1]
\State \textbf{Require} A 4D tensor $pcd4d$ containing point cloud data, resolution/voxel size, height and width of the BEV map, depth size for voxelization.
\State \textbf{Ensure} A 2D tensor $M$ representing a binary BEV map.
\State $x \gets pcd4d[0,:]$
\State $y \gets pcd4d[1,:]$
\State $z \gets pcd4d[2,:]$
\State $V \gets Zeros(width, height, depth size)$
\For{each point $i$}
    \State $voxelx \gets \lfloor \frac{x[i]}{resolution} + \frac{width}{2} \rfloor$
    \State $voxely \gets \lfloor \frac{y[i]}{resolution} + \frac{height}{2} \rfloor$
    \State $voxelz \gets \lfloor \frac{z[i]}{resolution} + \frac{deptsize}{2} \rfloor$
    \State $V[voxelx, voxely, voxelz]$$ \gets$ $MAX(V[voxelx, voxely, voxelz], z[i])$
\EndFor
\State $M \gets \text{max}(V, \text{axis}=2)$
\State $M \gets \text{where}(M > threshold, 1, 0)$
\State \textbf{return} $M$
\end{algorithmic}
\label{algo1}
\end{algorithm}

Given the current scan \( S_t \) with $1000$x$1000$ pixels in spacial dimension, a sequence of scans of the past $t-N$ to the present $t$ time steps are stacked along a new dimension which serves as the time dimension for each network’s input. The RACECAR dataset already has all LiDAR sweeps pre-aligned to the viewpoint of the ego vehicle. Now to learn the relationship across space and time, 2D convolutions can be applied to the BEV maps with the initial number of input channels corresponding to the number of input scans stacked along the time dimension.

\begin{figure}[t]
    \centering
    \includegraphics[width=1\linewidth]{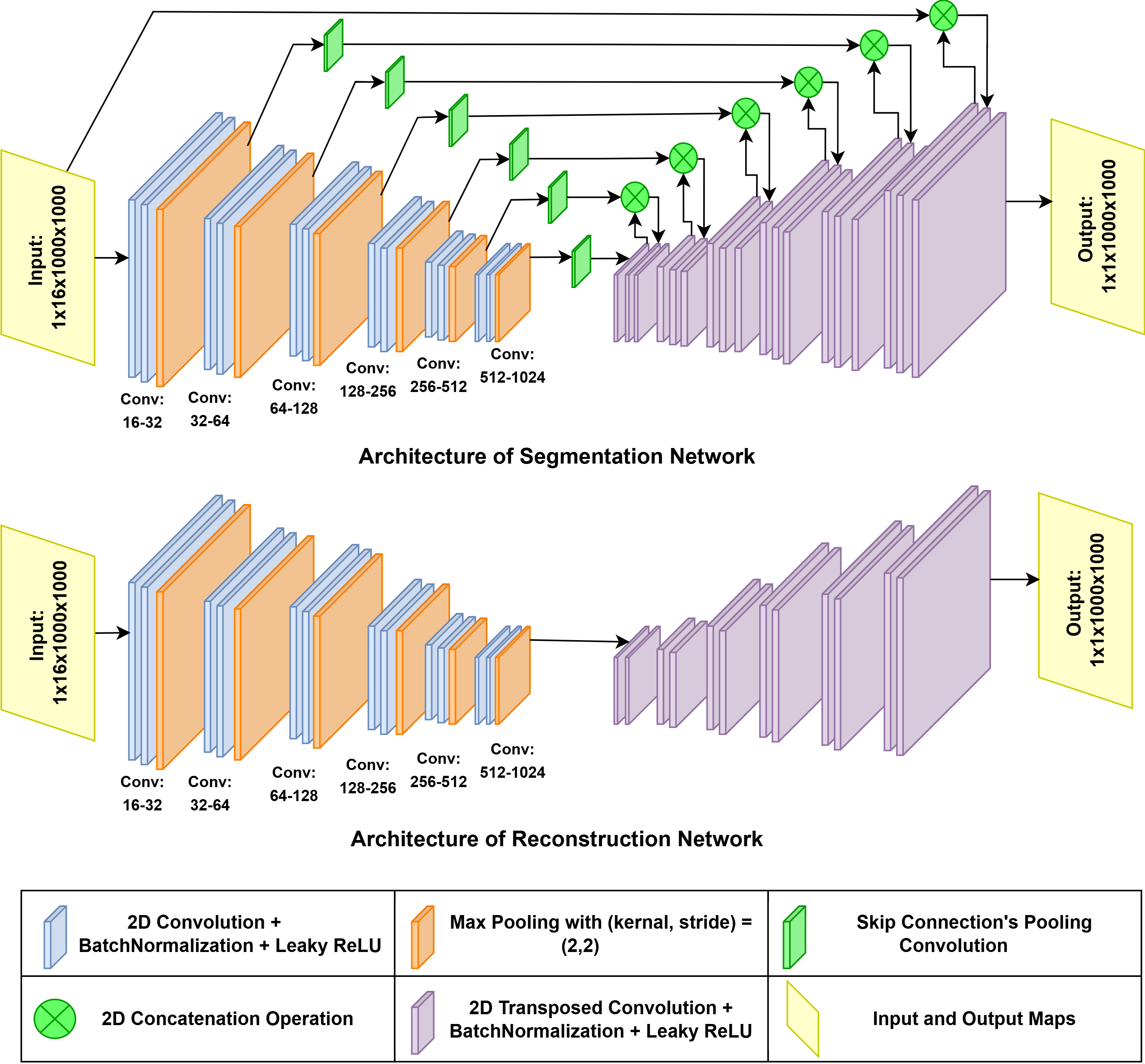}
    \caption{Architecture of Parallel Perception Network.}
    \label{fig:PPNarch}
\end{figure}

In Parallel Perception Network, input to each parallel network is a sequence of 2D binary BEV maps which can be considered as pseudo-images representing the environment states. The spatial and temporal features captured by 2D convolution allow the networks to learn patterns and changes in the input over time. Both the Segmentation Network and the Reconstruction Network in PPN consist of two parts, refer to Fig \ref{fig:PPNarch}:
\begin{enumerate}
\item \textbf{Encoder:} The encoder captures spatial hierarchies by applying a series of convolutions followed by batch normalization and Leaky ReLU activation that increases the channel depth while reducing spatial resolution, a single series is repeated to form a block of convolution with two layers. A max-pooling layer follows each convolution block. Batch normalization makes the network training stable and faster, while Leaky ReLU avoids the problem of dead neurons in normal ReLU activation functions.
\item \textbf{Decoder:} The decoder then reconstructs the spatial dimensions through blocks of transposed convolutions. This up-sampling process begins with the deepest level features which apply a series of deconvolution blocks consisting of transposed convolutional layers followed by batch normalization and Leaky ReLU activation as can be observed in Fig \ref{fig:PPNarch}.
\end{enumerate}

The skip connections which distinguish the two networks play a crucial role by enabling the concatenation of encoded features from different layers of the encoder with the corresponding decoder layer while upsampling. This feature fusion mechanism is incorporated into the segmentation network to retain the captured input representation information from the encoder to the decoder.

The first convolution to the input is a pseudo-1D convolution with a kernel size of \(T\)x$1$x$1$, where \(T\) represents the time dimension channel of the input. This captures the features from each pixel along the sequence of BEV maps from past to present time steps. Adding skip connections to this results in an accurate and detailed reconstruction of scenes segmented along space-time dimensions by PPN's segmentation network, without any training required specifically for constructing the segmentations. The rest of the following convolutions in the encoder are regular 2D convolutions. The decoder blocks apply 2D transposed convolution on the final encoded features of the encoder block, the transposed features are concatenated with the corresponding encoder layer skip connection’s pooling convolution features. This pyramid-shaped architecture can compute feature hierarchy along space-time dimensions utilizing only 2D convolutions making it highly efficient.

The reconstruction network, proposed as an additional network does independent learning in parallel and is targeted towards intelligent agents which require multi-network architecture for a simultaneous gain of knowledge from multiple perspectives for multiple tasks at once. To implement our proposed parallel neural network computing baseline architecture, PPN's reconstruction network follows the same convolution architecture as the segmentation network but lacks skip connections. Due to this, the encoded features are not concatenated at any layer while decoding making this network require training to reconstruct the input scenes.

The training approach of PPN's reconstruction network is designed to optimize the network’s parameters for accurate scene evolution reconstruction based on LiDAR scan images. The edge preservation is combined as a loss function by utilizing Canny Edge Detection with MSE loss and SmoothL1 loss. The resulting function with a linear combination of the weighted sums of SmoothL1 loss and MSE loss with corresponding edge detection losses, named Mean Square Smooth Canny Edge (MSSCE) loss provides a balance between robustness to outlines and precision in regression while preserving sharpness using (\ref{lmssce}).

Equation (\ref{eq1}) represents the MSE and SmoothL1 loss functions, while (\ref{ledge}) represents the edge-preserving loss function.
\begin{equation}
    \textit{L}_{\text{MSE}} = \frac{1}{N} \sum_{i=1}^{N} (y_i - \hat{y}_i)^2
    \label{eq1}
\end{equation}
\[\textit{L}_{\text{SmoothL1}} = 
    \begin{cases} 
    \frac{(0.5(y_i - \hat{y}_i))^2}{\beta}, & \text{if } |y_i - \hat{y}_i| < \beta \\
    |y_i - \hat{y}_i| - 0.5*\beta, & \text{otherwise}
    \end{cases}\]
\begin{equation}
    \textit{L}_{\text{Edge-Preserving}} = \frac{1}{N} \sum_{i=1}^{N} \left| \text{C}(y_i) - \text{C}(\hat{y}_i) \right|
    \label{ledge}
\end{equation}
Where: \(N\) is the number of pixels in the image, \(y_i\) is the ground truth value at pixel \(i\), \(\hat{y}_i\) is the predicted value at pixel \(i\), $\beta$ is a hyperparameter that controls the transition point between the two regions of SmoothL1 loss function, and \(\text{C}(x)\) represents the Canny edge detection applied to the image \(x\). This loss function calculates the absolute difference between the Canny edge detections of the ground truth and predicted images, averaged over all pixels.

Considering (\ref{eq1}) and (\ref{ledge}),
\begin{enumerate}
\item The MSE Loss + Edge Preserving Loss can be derived as:
\begin{align}
    \textit{L}_{\text{MSE+Canny}} &= \frac{1}{N} \sum_{i=1}^{N} \Big( \lambda (y_i - \hat{y}_i)^2 \notag \\
    &\quad + (1 - \lambda) \left| \text{C}(y_i) - \text{C}(\hat{y}_i) \right| \Big)
\label{lossf1}
\end{align}
\item And SmoothL1 Loss + Edge Preserving Loss can be derived as:
\begin{align}
    \textit{L}_{\text{SmoothL1+Canny}} &= \frac{1}{N} \sum_{i=1}^{N} \Big( \lambda (\textit{SmoothL1} ( y_i - \hat{y}_i )) \notag \\
    &\quad + (1 - \lambda) \left| \text{C}(y_i) - \text{C}(\hat{y}_i) \right| \Big)
\label{lossf2}
\end{align}
\end{enumerate}

The loss functions represented by (\ref{lossf1}) and (\ref{lossf2}) integrate Mean Squared Error (MSE) and SmoothL1 losses respectively, with corresponding edge preserving terms. Here, \(N\) is the total number of pixels in the image, \(y_i\) and \(\hat{y}_i\) are the ground truth and predicted values at the \(i\)-th pixel respectively. The term \((y_i - \hat{y}_i)^2\) in (\ref{lossf1}) calculates the squared difference between the ground truth and predicted values, representing the MSE loss. Conversely, the ${SmoothL1}(y_i - \hat{y}_i)$ term in (\ref{lossf2}) calculates the SmoothL1 loss between the ground truth and predicted values. The \(\lambda\) parameter is a weighting factor that balances the MSE and SmoothL1 terms with the edge-preserving term, \(\left| \text{C}(y_i) - \text{C}(\hat{y}_i) \right|\), which computes the absolute difference between the Canny edge detections of the ground truth and predicted images. By adjusting \(\lambda\), the trade-off between preserving edges and ensuring pixel-wise accuracy in the predicted image can be controlled.

Finally, from (\ref{lossf1}) and (\ref{lossf2}) the proposed Mean Square Smooth Canny Edge Loss is defined as:
\begin{equation}
    \textit{L}_{\text{MSSCE}} = \textit{L}_{\text{MSE+Canny}} + \textit{L}_{\text{SmoothL1+Canny}}
\label{lmssce}
\end{equation}
Upon successful training and deployment, PPN demonstrates a high degree of accuracy in segmenting the scene evolution and reconstruction of the scene segmentation in parallel from sequential data, showing the effectiveness of the combined loss function in calibrating the network in understanding the input sequence and reconstructing the output.
\section{Experiments and Evaluation}
In this section, utilizing the \textbf{RACECAR} dataset’s LiDAR sensor data, which is in nuScenes format and spans $11$ racing scenarios from fully self-driving racecars going at speeds up to $274 kmph$, we train the PPN model's networks on a set of LiDAR sweeps from one race scenario, the PoliMove team's Multi-Agent Slow on LVMS racetrack, containing $7,150$ sweeps and evaluate its inference time performance against a sequential model setup.
\subsection{Implementation Details}
In our implementation, the PPN model crops point cloud data to reside within a region defined by a $1000$x$1000$ grid, corresponding to the XY plane resulting in a binary BEV map that captures the environment around the ego vehicle. The conversion process involves mapping each point in point cloud data to its corresponding pixel location on the BEV map. Points falling outside the boundaries are disregarded. The map is updated using MAX pooling operation between existing map values and the Z value of the points that correspond to the same pixel location from the point cloud. This retains elevations at each pixel location in the 2D representation of 3D data.

\begin{figure}[t]
    \centering
    \includegraphics[width=0.75\linewidth]{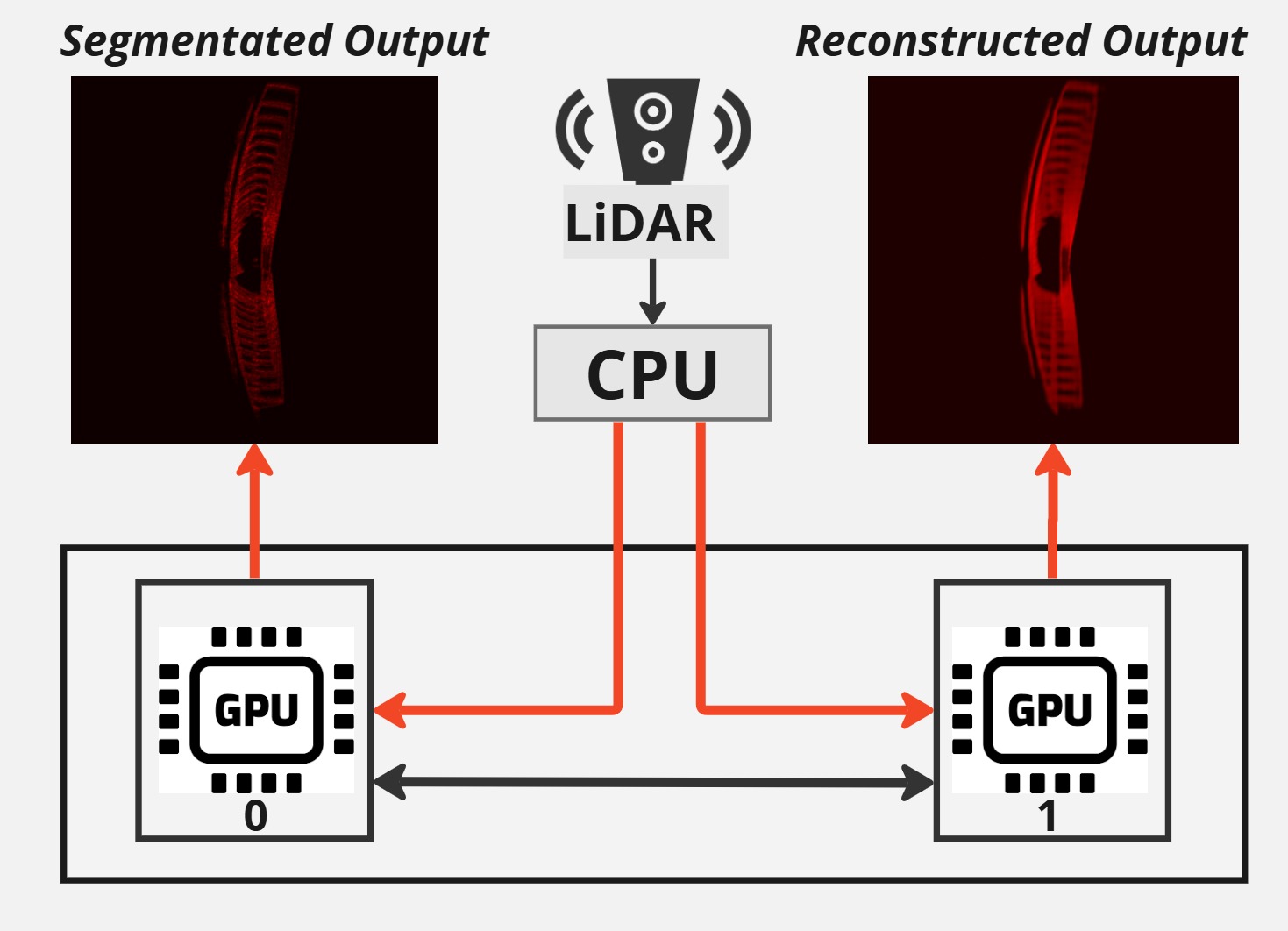}
    \caption{PPN model's experimental setup on parallel accelerated hardware.}
    \label{fig:parallelaccHW}
\end{figure}

To capture the temporal dynamics of the environment we ingest a set of $15$ consecutive past scans in addition to the current $t^{th}$ scan as the present scan, these scans span from $(t-15)^{th}$ to $t^{th}$ time frames. We choose these numbers as per the LiDAR sensor configuration of vehicles used to record sensor data, the RACECAR’s sensor provides $10-30$ frames per second. Looking at the mathematics for a racecar that travels at speeds of $200 kmph$ which translates to $55$ meters per second, this is a substantial distance covered in just a second. Choosing time frames to capture historical context within half a second for our network provides an appropriate temporal window in high-speed racing applications. Our system has $2$ CUDA-enabled NVIDIA T4 GPUs (see Fig \ref{fig:parallelaccHW}) with $16$ GB graphics memory each.
\subsection{Training Details}
Due to the lack of hand-labelled annotations in the RACECAR dataset, we employ the segmentation network's output as ground truths for the corresponding input sequence to train the reconstruction network to demonstrate said parallel neural computing baseline. Since the LiDAR scans converted into BEV maps capture the spatial distribution of points representing the environment and track layout, a loss function that captures the spatial overlap and structural integrity is essential. We experimented with Intersection over Union (IoU) as a loss function to ensure that the prediction and ground truth match closely by maximizing their overlap. However, opting for edge preservation using the Canny operator with a combined SmoothL1 and MSE loss function (Mean Square Smooth Canny Edge loss) to preserve the structural integrity resulted in sharper and more accurate reconstructions.

We also benchmark the prediction accuracy of our segmentation network by training and validating this network's modified number of output channels against future BEV maps as ground truths targeting the scene evolution over time $(t+d)^{}$ to $(t+d+F)^{}$. Here, $d$ is the computation time to predict scene evolution over the next $F$ time frames. The future scans serve as a benchmark for this network’s prediction to quantify its accuracy through the combined SmoothL1 and MSE losses only. To ensure both networks learn from their ground truths, the training is optimized using the Adam optimizer, which handles sparse gradients leading to swift network convergence. The hyperparameters are set as follows: learning rate = $10^{-4}$, $\beta$ = $1$ and $\lambda$ = $0.85$.
\subsection{Results and Analysis}
\begin{figure}[t]
    \centering
    \includegraphics[width=0.75\linewidth]{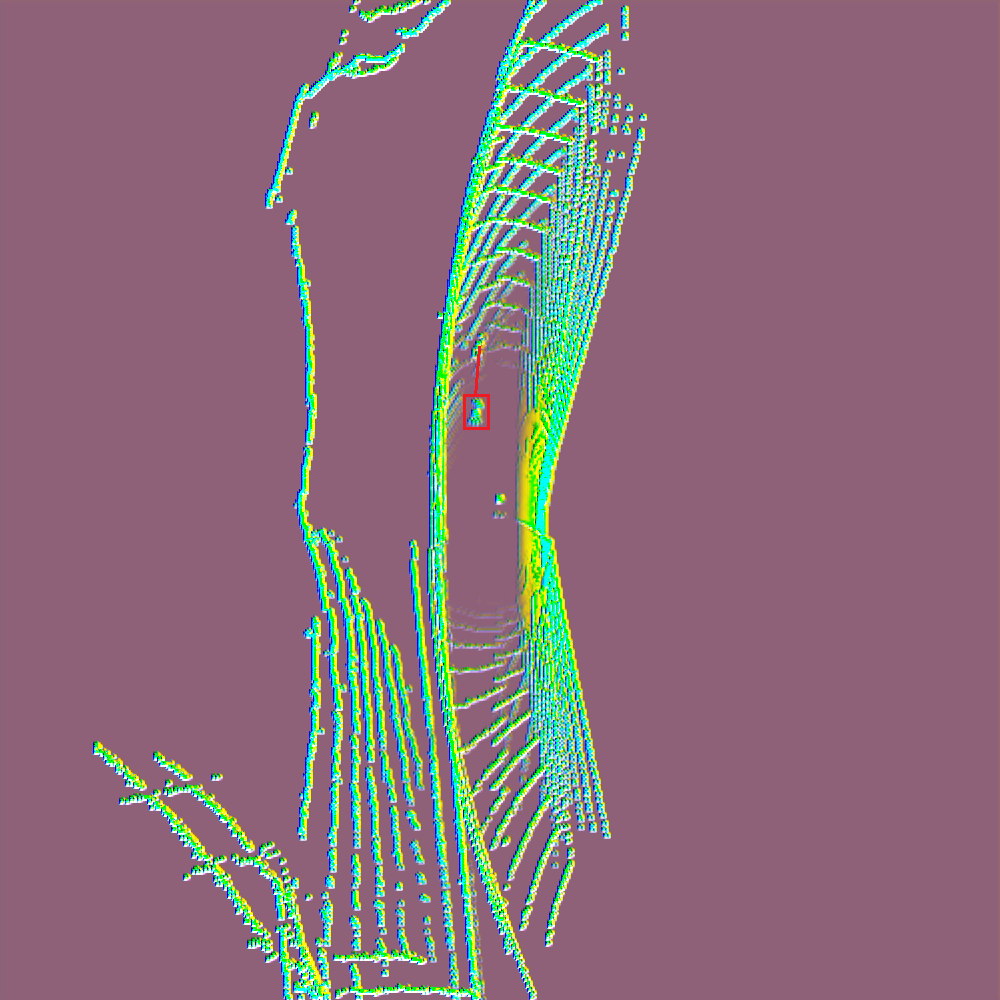}
    \caption{RGB image of segmented output map with hand-annotated motion information, the red box shows the current position of the vehicle and the red line shows its motion from $(t-15)^{th}$ to $t^{th}$ time frame.}
    \label{fig:rgbmap_motion}
\end{figure}
\begin{figure}
    \centering
    \includegraphics[width=0.75\linewidth]{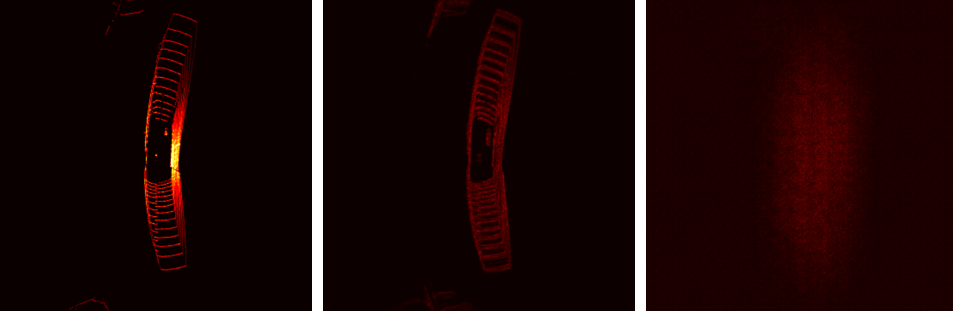}
    \caption{PPN model’s input and outputs without training. Left to Right: Current scan from Input Sequence, Segmentation network output, Reconstruction network output.}
    \label{fig:notrain_ppnout}
\end{figure}
To show that our model's segmentation network accurately segments the scene along space-time dimensions capturing a brief history of motion of other racecars on the track, without any training required, we modify the network's input and output layers to get an RGB image of the output map. The input layer is modified to have 2 channels corresponding to BEV maps at $(t-15)^{th}$ and $t^{th}$ time frames. This helps us visualise a racecar's positions at the beginning and end of the input sequence capturing its travel history. We alter the output layer to have 3 channels and replace its activation with a tanh activation function followed by rescaling the output values to a range of $(0, 255)$. PIL (Python Imaging Library) is then used to convert the 3-channel output into a single RBG image of size $1000$x$1000$ pixels. Fig \ref{fig:rgbmap_motion} shows the resulting RGB image with hand-annotated motion information for demonstration. Fig \ref{fig:notrain_ppnout} shows the PPN model's current input scan and outputs from each parallel network without training. The output of the untrained reconstruction network is a blank image resulting from the removal of skip connections.
\begin{table}
\begin{tabular}{|p{2.5cm}||p{2cm}|p{1.25cm}|p{1.25cm}|}
 \hline
 PPN's Networks& \multicolumn{3}{c|}{Accuracy after training with loss functions} \\ \cline{2-4}& MSE+SmoothL1 &IoU &MSSCE\\
 \hline
 Segmentation Net   &98.2\%    &-   &-\\
 Reconstruction Net   &-  &99\%   &61\%\\
 \hline
\end{tabular}
\caption{Post-Training Accuracy of Segmentation and Reconstruction Networks.}
\label{table1}
\end{table}

\begin{figure}[t]
    \centering
    \includegraphics[width=1\linewidth]{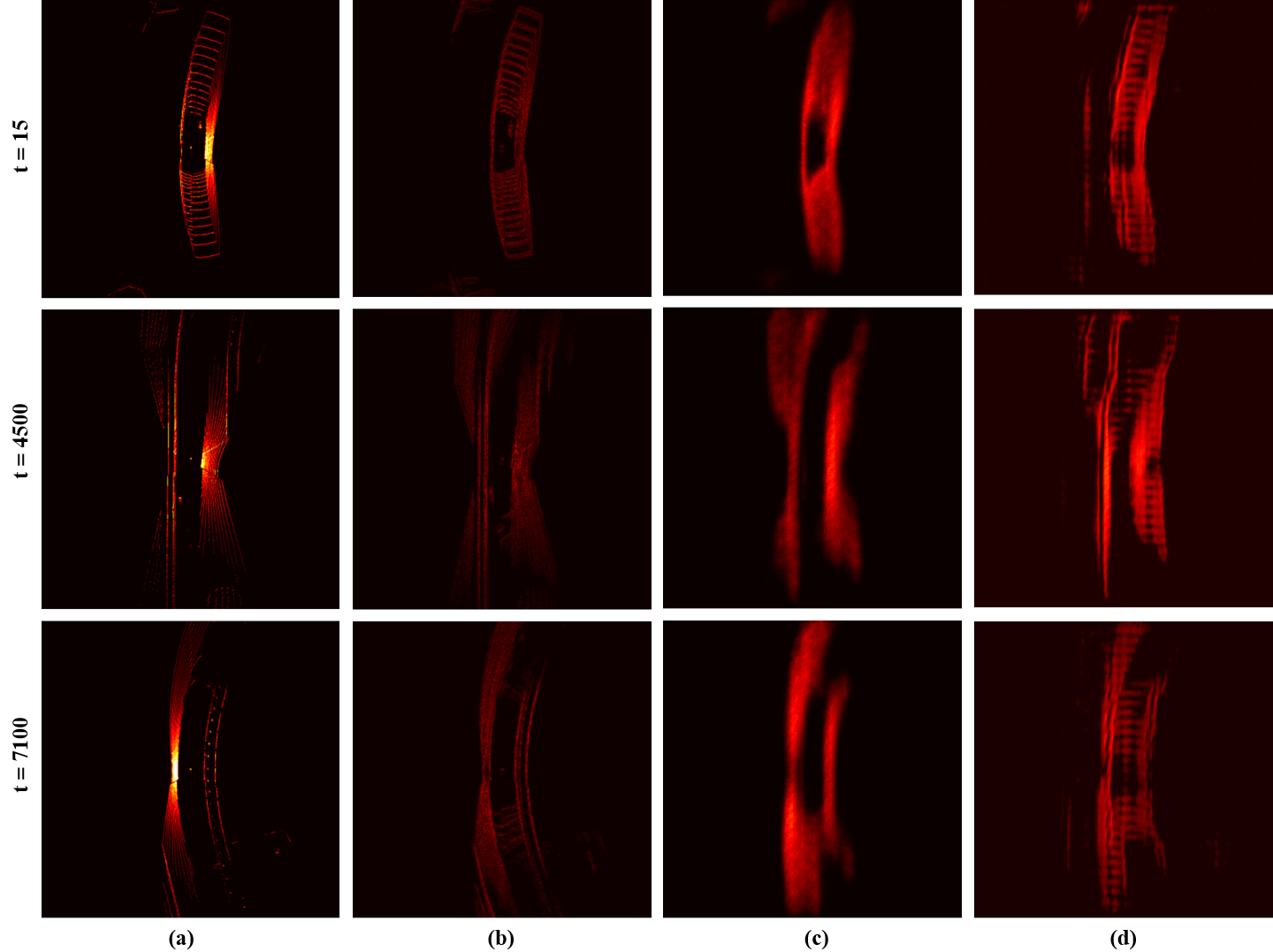}
    \caption{Qualitative Results of trained PPN model. \textbf{Top to bottom:} Inputs and outputs at various time frames corresponding to $t = 15$, $t = 4500$ and $t = 7100$. \textbf{Left to Right:}  (a) Current scans from input sequences at different time steps, (b) Segmentation Network outputs i.e. scenes segmented along space-time dimensions, (c) Reconstruction Network outputs i.e. scene reconstructed trained with IoU loss, and (d) Reconstruction Network outputs i.e. scene reconstructed trained with MSSCE loss.}
    \label{fig:quantres-ppnmodel}
\end{figure}

Table \ref{table1} summarizes the post-training prediction accuracy of the segmentation and reconstruction networks, trained for $700$ iterations with two loss functions. Fig \ref{fig:quantres-ppnmodel} shows the qualitative results of our PPN model with trained networks. As discussed, the MSSCE loss (Fig 7d) does a better job at training the reconstruction network resulting in a sharper reconstruction than the blurred ones with the IoU loss (Fig 7c).
\subsection{Performance Evaluation}
\begin{table}
\begin{tabular}{|p{0.2cm}|p{4cm}||p{3cm}|}
 \hline
 &Configuration &Inference cycle speeds in seconds (min and max)\\
 \hline
 1. &Sequential, segmentation done first followed by reconstruction, with 1 GPU for both networks. &\hspace{5pt}0.162\hspace{30pt}0.205\\
 2. &Parallel, segmentation and reconstruction done simultaneously, with seperate GPUs for each network. &\hspace{5pt}0.075\hspace{30pt}0.091\\
 \hline
\end{tabular}
\caption{PPN model's infernece speed comparision.}
\label{table2}
\end{table}
We list the model inference times measured across multiple runs for different configurations in Table \ref{table2}. The observed effect of running each network on separate hardware demonstrates the significant advantage of exploiting true hardware-enabled parallelism for multi-network architectures. These measurements were conducted on a system with NVIDIA T4 GPUs and reveal a speedup of at least two times for the parallel configuration compared to the sequential one.

\subsection{Comprehensive Comparison}
Unlike approaches such as MotionNet \cite{wu2020motionnet}, InsMOS \cite{wang2023insmos}, LookOut \cite{cui2021lookout}, and many more which perform joint perception and prediction tasks in a single network pipeline or by fusing multiple sensor data, models built on the proposed architecture would process data from multiple sensors, such as cameras, LiDARs, radar, GNSS, using independent neural networks, each running on its own GPU. Here, the model would understand the scenes and environment from different sensor data perspectives simultaneously where each network is specialized in feature learning from each type of input data. Table \ref{table3} highlights the advantages and limitations compared to existing methods.
\begin{table}
\begin{tabular}{|p{1.3cm}|p{3cm}|p{3cm}|}
 \hline
 Method &Advantage &Limitation\\
 \hline
 PPN(Ours) &Hardware enabled parallelism for faster processing and learning distinct features from each type of sensor separately and within a single inference cycle, providing a richer representation of the environment and performance scalability. The use of Canny Edge Detection in loss function helps preserve sharp features. &Requires multiple GPU hardware, current implementation is limited to LiDAR sensor data and there could be synchronization overhead between networks processing different types of sensor data.\\
 Single Pipeline and sensor data fusion &Simpler architecture, lower hardware requirements, unified feature learning. &Limited performance scalability, higher processing latency. \\
 \hline
\end{tabular}
\caption{Comparision with existing methods}
\label{table3}
\end{table}

\section{Conclusion}
In conclusion, we present a novel baseline architecture for parallel computing of neural networks on accelerated hardware. The presented model PPN can do parallel perception and predictions by directly feeding on LiDAR point clouds and converting them into a binary BEV map representation to feed to each network. The resulting 2x speedup in model inference time, when compared to a sequential setup, shows the true potential of enabling parallel acceleration for multi-network model architectures for complex and sophisticated intelligent agents, especially a high-speed autonomous racing agent.

\bibliography{myref.bib}{}
\bibliographystyle{plain}
\end{document}